\documentclass[letterpaper, 10 pt, conference]{ieeeconf}
\IEEEoverridecommandlockouts                              

\let\labelindent\relax
\usepackage[loadonly]{enumitem}
\usepackage{cite}
\usepackage{amsmath,amssymb,amsfonts}
\usepackage{algorithmicx}
\usepackage{graphicx}
\usepackage{textcomp}
\usepackage{xcolor}
\usepackage{booktabs}
\usepackage{pifont}

\usepackage{multirow}
\usepackage{textcomp}
\usepackage{pgfplots}
\pgfplotsset{compat=1.9}
\usepackage[hidelinks]{hyperref}  
\usepackage[capitalise,noabbrev]{cleveref}  

\usepackage[table]{xcolor} 
\definecolor{lightgray}{gray}{0.9}
\usepackage{tabularx}

\begin{document}

\title{BronchoTop: Bronchoscopy Navigation via RGB-Only Topological Localization \\

\author{Clara Tomasini$^{*}$, Ana Cristina Murillo$^{*}$ and Luis Riazuelo$^{*}$}
\thanks{$^{*}$ This work was conducted at DIIS - I3A - Universidad de Zaragoza, Spain. {\tt\small ctomasini@unizar.es}}%
\thanks{This work was partially funded by project T45\_23R, by grants PID2022-139615OB-I00, AIA2025-163563-C31, PID2024-159284NB-I00,  funded by MCIN/AEI/10.13039/501100011033 and ERDF.}
}

\maketitle
\begin{abstract}
Accurate localization of the bronchoscope within the bronchial tree is essential for clinicians to be able to reach target lesions, perform biopsies and avoid misidentification of airway segments during diagnostic and therapeutic procedures. However, existing navigation systems typically rely on patient-specific CT scans or additional external sensors, increasing cost, setup time and patient radiation exposure. This work presents BronchoTop, a real-time, RGB-only framework for topological bronchoscopy localization that eliminates the need for patient-specific data. BronchoTop estimates scope location relative to a generic airway model through four modules: lumen detection and tracking, lumen-branch label association, probabilistic scope location estimation, and switch verification. By using only standard bronchoscopy video input, BronchoTop provides practical, real-time navigational assistance to physicians. Evaluation on phantom, simulated and real data demonstrates state-of-the-art accuracy, improving existing approaches performance by over 20\% on real bronchoscopy sequences. 
BronchoTop is the first published framework\footnote{https://sites.google.com/unizar.es/bronchotop} including both the localization algorithms as well as all the real data used, together with code to generate additional simulations, encouraging and facilitating further developments and benchmarking. The results highlight BronchoTop's potential to enhance procedural safety, efficiency and accessibility in clinical and robotic bronchoscopy.
\end{abstract}
   
\section{Introduction}
Video bronchoscopy is a key diagnostic and therapeutic tool in respiratory medicine. It enables clinicians to visually explore the bronchial tree with a camera (bronchoscope) to detect potential lesions or tumors, biopsy them, or retrieve foreign objects.

\begin{figure}[tb]
    \centering
    \includegraphics[width=1\linewidth]{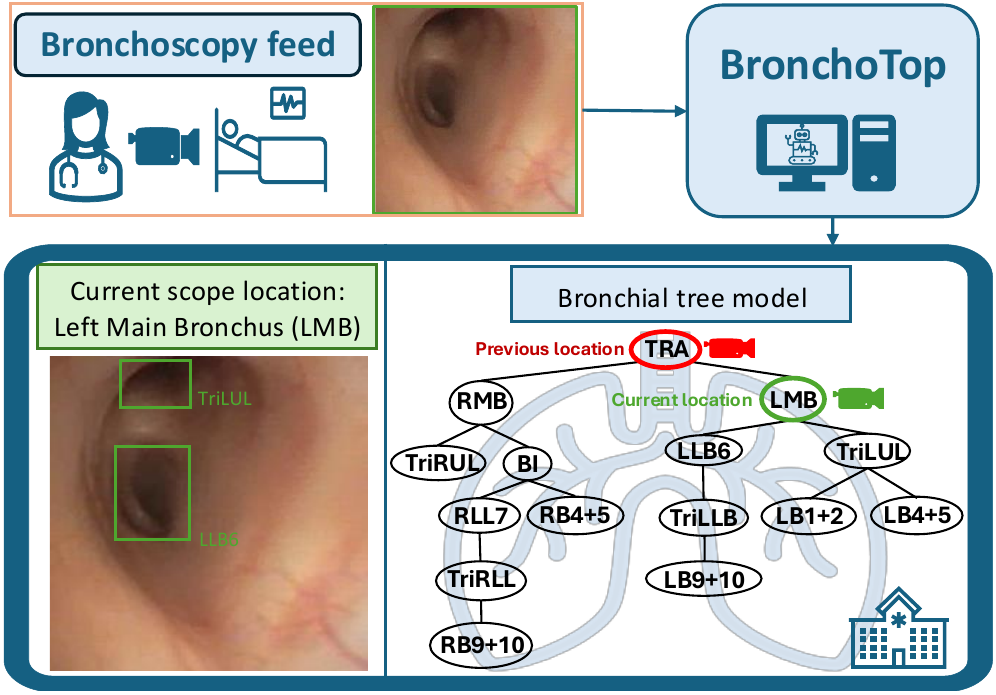}
    \caption{Proposed topological localization system proposed. From the RGB bronchoscopy feed, BronchoTop localizes the scope within a clinically used, generic tree model of the airway, common between patients, shown on the right. Nodes, like TRA (Trachea), LMB (Left Main Bronchus), RMB (Right Main Bronchus), represent airway branches and edges encode parent-child relationships. The highlighted node in green indicates the predicted current location (LMB in this case). No additional input or setup is needed. }
    \label{fig:intro}
\end{figure}

During the bronchoscopy, clinicians must mentally track the bronchoscope's position through the complex tree-like structure of the airway (Figure~\ref{fig:intro}) to locate potential regions of interest as they navigate through the airway. Misidentifying  airway segments can lead to longer procedures and missed lesions, and navigational assistance systems could therefore prove essential to improve accuracy, efficiency~\cite{shinagawa2007virtual}, and clinical outcomes~\cite{merritt2008image,memoli2012meta}.

Several techniques have been developed to address these challenges. Current navigation methods, such as Electromagnetic Navigation Bronchoscopy (ENB)~\cite{reynisson2014navigated} and Virtual Navigation Bronchoscopy (VNB)~\cite{ferguson2005virtual}, typically rely on pre-operative CT scans to obtain a 3D reconstruction of the patient's airway, and external sensors (ENB) or image registration~\cite{mori2004new} (VNB) to track the bronchoscope through the 3D reconstruction during the procedure.  To improve VNB performance, various methods combine depth inference with CT registration~\cite{shen2019context,shen2015robust,banach2021visually}, or include scope motion~\cite{tian2024dd,tian2024pans} or additional semantic information from the image~\cite{wang2023anatomy,zang2023bronchoscopic,shen2017branch,tian2024pans} in an effort to improve registration. 
Other approaches propose to use Simultaneous Localization And Mapping (SLAM)~\cite{wang2019visual,wang2020visual} to obtain a 3D reconstruction of the explored airway and register it to the patient specific CT scan, or use CNNs~\cite{borrego2023bronchopose} and NeRFs~\cite{mildenhall2020nerf,zhu2024bronchoscopic} to estimate camera pose without registration to CT. While effective, these methods require additional hardware and patient-specific CT scans that increase patient radiation exposure and setup time. 

However, precise metric localization is often unnecessary. Instead, topological localization with regard to a generic, clinically used airway model (Figure~\ref{fig:intro}) provides sufficient spatial context for navigational guidance and paves the way for future autonomous procedures.

\begin{figure*}[tb]
    \centering
    \includegraphics[width=0.9\linewidth]{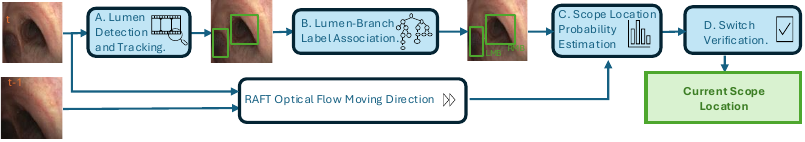}
    \caption{\textbf{Proposed BronchoTop pipeline.} Our approach consists of 4 modules: A) Lumen detection and tracking, B) Lumen-Branch Label Association with regards to the bronchial tree model, C) Scope Location Probability Estimation, and D) Switch Verification to guarantee the detected location switch is real and not due to brief detection or tracking failures. Module C relies on the tracked labeled lumens as well as the camera moving direction, obtained from RAFT~\cite{teed2020raft} optical flow estimation.}
    \label{fig:pipeline}
\end{figure*}

This work introduces BronchoTop, an online, RGB-only topological localization framework that provides real-time bronchoscopy navigation assistance without requiring patient-specific CT scans, or additional hardware. As illustrated in Figure~\ref{fig:pipeline}, our method estimates the scope location within a generic bronchial tree model shared across patients and relies only on the RGB bronchoscope video feed. Our evaluation shows BronchoTop's significant improvements compared to existing approaches when run on real data without requiring any labeled topological location data. 
BronchoTop relies largely on training-free structural tracking, with only one module requiring light easily obtainable supervision to estimate image similarity.

In summary, our main contributions are the following:
\begin{itemize}
    \item We introduce a novel, \textbf{image-only approach for online topological localization in bronchoscopy}. Comprising of four modules (lumen detection and tracking, lumen-branch label association, scope location probability estimation, and location switch verification)  our method achieves state-of-the-art performance on real bronchoscopy recordings. 
 
    \item We present the first publicly available \textbf{dataset for bronchoscopy topological localization}, releasing annotated real procedures and synthetic sequences to foster reproducibility and future research in this data-scarce domain.
    \item We release our complete \textbf{framework} and trained models, enabling future benchmarking and the generation of new simulated data.
\end{itemize}

\section{Related Work}
Automatically localizing the scope as it traverses the organ being explored is a long studied, challenging problem due to the complex non-sequential structure of the bronchial tree. 

Conventional methods rely on auxiliary data and sensors, such as ENB~\cite{franz2014electromagnetic} and VNB~\cite{rai2008combined}, while image-only approaches often build dense 3D models using SLAM techniques~\cite{wang2019visual,wang2020visual}, or rely on CNN~\cite{borrego2023bronchopose} and NeRFs~\cite{mildenhall2020nerf,zhu2024bronchoscopic} to obtain the camera pose. 
However, accurate camera pose estimationis not always necessary, and topological localization often suffices for applications navigation assistance. 

\paragraph*{Location recognition}
Recent works propose to localize the bronchoscope by classifying each frame along the sequence as one of the possible locations, or places, within the bronchial tree. 
For example, ~\cite{yoo2021deep} and ~\cite{chen2023distinguishing} apply supervised CNN models, such as EfficientNet~\cite{tan2019efficientnet}, ResNet50v2~\cite{he2016identity} and DenseNet~\cite{huang2017densely} for frame-wise classification in the initial levels of the tree. Similarly, ~\cite{liu2024artificial} detects airway lumens using YOLOv7~\cite{wang2023yolov7}to deduce frame location based on the structure of the bronchial tree.
However, these approaches classify each frame individually and do not take into account the temporal aspect of a bronchoscopy procedure and its influence on the localization of the scope. Additionally, neither the trained models nor the large sets of labeled real bronchoscopy images used for training are public up to our knowledge. 

\paragraph*{Topological localization with temporal consistency}
To incorporate temporal dynamics, recent approaches track visible lumens to build exploration trees. For lumen detection, methods range from fitting ellipses to visible lumens~\cite{esteban2016stable} or using image gradients to detect lumen centers~\cite{sanchez2013line}, to depth-based strategies where depth-maps generate pseudo-labels~\cite{keuth2023weakly} for Lite R-ASPP segmentation~\cite{howard2019searching} or are thresholded to obtain lumen segmentations~\cite{wang2021depth,wang2022bronchial} after being generated using CycleGAN~\cite{zhu2017unpaired}.
By tracking these lumens, several works~\cite{wang2021depth,wang2022bronchial,gil2020intraoperative,sanchez2016navigation,esteban2016stable} dynamically generate a generic exploration tree. However, these relative trees lack direct correspondence to clinical anatomy, meaning they can identify branch depth but cannot differentiate specific nodes at each level, Right Main Bronchus (RMB) and Left Main Bronchus (LMB). Other methods~\cite{tian2024bronchotrack,sanchez2016navigation} assign absolute labels by registering the exploration tree to a patient-specific pre-operative CT scan, as is already done for Virtual Bronchoscopic Navigation. While effective, the reliance on patient-specific CT scan limits their direct clinical applicability and introduces radiation exposure and setup overhead.

\paragraph*{Topological localization within anatomical models}
To overcome the need for patient-specific CT scans and generate exploration trees with clinical meaning, \cite{keuth2024airway} localizes the scope within a generic, inter-patient model of the airway, as the airway structure has been proven similar enough up to the fourth branch level to do so~\cite{smith2018human}. This method uses a CNN (trained on k-means pseudo-labels from depth-maps~\cite{keuth2023weakly}) to compute location likelihoods. A Hidden Markov Model (HMM) then applies anatomical constraints to estimate the explored path. However, this approach is defined as an offline localization pipeline, and could not be used online, for navigational guidance during the procedure, due to the constraints and formulation of the HMM. For each sequence, the locations of the first and last frames (beginning and end locations) are required to compute each frame's location.
This method shows good results in phantom data, but it has not been tested in real data. 


Our approach, BronchoTop, proposes an online  topological localization method that relies exclusively on the RGB images captured by the bronchoscope. The localization is done within the same generic model of the bronchial tree used in~\cite{keuth2024airway}, removing the need for patient-specific CT scan, challenging registration techniques. BronchoTop allows  for simplified and homogeneous identification of previously detected regions of interest in the case of follow-up procedures, and presents promising results on our released real procedure dataset.

\section{Method}
\label{sec:method}
We present BronchoTop, a novel pipeline for online topological bronchoscope localization with regard to a generic topological model of the bronchial tree. This clinically used airway topology model consists of 15 classes over 6 levels. An essential aspect of our approach is that it uses only the image data (frames) from the scope and does not require any external sensor or patient data. 

For each input frame at instant $t$, our approach computes $P_{loc,t}(k)$, the probability of the scope being at node $k$ of the tree, for every possible $k$. From these probabilities, the most likely position of the scope $S_t$ is determined. Our pipeline consists of four steps (A.~\textit{Lumen Detection and Tracking}; B.~\textit{Lumen-Branch Label Association}; C.~\textit{Scope Location Probability Estimation}; D.~\textit{Switch Verification}), as illustrated in Figure~\ref{fig:pipeline} and described in the following subsections. 

\subsection{Lumen detection and tracking.} \label{sec:tracking} 
The input to this module are the RGB frames. Each frame is first converted to grayscale, then K-means clustering (with $K=5$) is applied to detect the visible lumens. As the lumens regions correspond to the furthest points from the camera, they generally form the cluster of the darkest pixels, with the lowest intensity. The detected lumens are then passed to an IoU-based multi-object tracker to track lumens across various frames. The IoU tracker used considers two detections as matched when the Intersection-over-Union (IoU) between their two bounding boxes is over 30\%, as used in~\cite{tian2024bronchotrack}. A tracked lumen is removed from the tracker if it is not detected for $M_{track} = 5$ consecutive frames. Tracking makes the detection and overall pipeline more robust to noise, inherently present in bronchoscopy frames due to varying lighting conditions and possible presence of liquids. The output of this module is a list of all lumens tracked at a given time t, with corresponding bounding box coordinates.

\subsection{Lumen-Branch Label Association.} \label{sec:ba}

\begin{figure}[tb]
    \centering
    \includegraphics[width=1\linewidth]{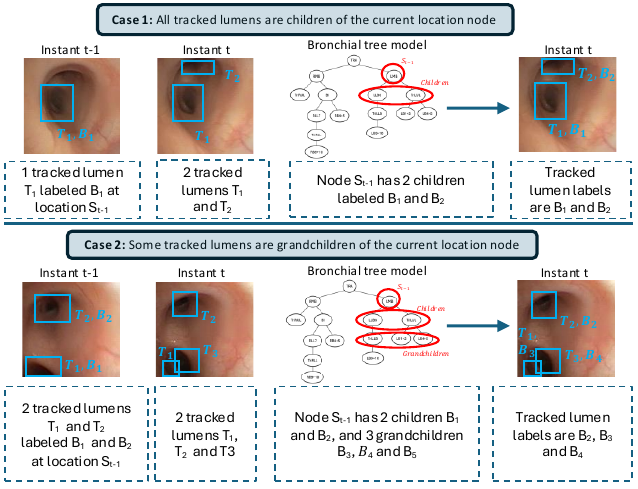}
    \caption{\textbf{Branch Association module} for two different cases. 1) The number of tracked lumens in the current frame is lower or equal to the number of children of location node $S_{t-1}$. All the lumens visible are children of $S_{t-1}$. 2) The number of tracked lumens in the current frame is greater than the number of children of location node $S_{t-1}$. Lumens tracked are children and grandchildren of $S_{t-1}$.}
    \label{fig:ba}
\end{figure}

This module's input is the list of tracked lumens at time $t$. At this stage,
each tracked lumen is assigned a label from the bronchial tree model. This label association is performed based on: 1) the location of the scope at instant $t-1$, 2) the lumens possibly visible at that location, and 3) the set of tracked lumens. 
At a specific node in the tree, the visible lumens can be children of that node, or children of the node's children (one level down, grandchildren). This fact defines the possible labels for the tracked lumens, included in the list of children and grandchildren of the previous scope location. Once the list of visible lumens is determined, the labels are assigned to tracked lumens  based on the bounding box center coordinates of each tracked lumen. 
Specifically, the tracked bounding boxes are ordered geometrically from left to right and assigned a label following the same order as in the bronchial tree model. This left-to-right assignment relies on the initial assumption that the bronchoscope enters the airway in an upright position. As the scope progresses and potentially rotates, its orientation might change, however this change is continuously captured and taken into account by the tracker.
To illustrate this process, Figure~\ref{fig:ba} shows the branch association process for two cases: when all visible lumens are children of the current node, and when visible lumens are a combination of children and grandchildren. 

\subsection{Scope Location Probability Estimation.} \label{sec:prob}
Given the labeled tracked lumens, the tree model and the camera moving direction, we estimate the scope location probability ${P_{loc,t}(k)}$ of being at node $k$ at instant $t$ using a recursive Bayesian update rule. Let $Z_t$ denote the observation (the number of active and inactive tracked lumens) and $d_t \in \{forward, backward\}$ denote the camera moving direction. The update rule is defined as:
\begin{equation}
    P_{loc,t}(k) = \eta P(Z_t|k)\sum_{j\in tree} P(k|j,d_t)P_{loc,t-1}(j)
\end{equation}
where $\eta$ is a normalization factor. As opposed to recent methods like~\cite{wang2022bronchial}, which relies on deterministic tracking of bronchial orifices to estimate branch levels, this probabilistic formulation allows for handling sporadic false positives and missed detections by maintaining a probability distribution over the entire generic anatomical tree. 

\paragraph{Transition probability $P_{k|j,d_t}$}

In bronchoscopy, the camera can move forward (towards deeper levels) or backward (towards the trachea) through the airway. The camera moving direction $d_t$ is determined by computing the optical flow between the previous and current frames. For this, we use widely established optical flow estimation model RAFT~\cite{teed2020raft}. To ensure real-time performance and bound the linear inference time, we limit the GRU updates to 12 iterations. 
The transition model couples the anatomical constraints from the bronchial tree model with $d_t$. The probability $P_{k|j,d_t}$ is non-zero only for anatomically valid transitions: at a transition point, moving to a child node when $d_t = forward$ or moving to a parent node when $d_t = backward$, or remaining at the current node $j$. By explicitly integrating the camera moving direction into the transition model $P_{k|j,d_t}$, our method restricts probability propagation to anatomically plausible location transitions, setting it apart from generic tree-building methods which are susceptible to topological drift if tracking temporarily fails.

\paragraph{Observation Likelihood $P(Z_t|k)$}

Each location in the bronchial tree can be represented by a branch, similar to a tunnel, and a branching point, where the branch separates. When moving from one location to the next, a critical variable is the number of lumens detected $Z_t$. 

\noindent In the case of forward movement, while the scope is in a branch, only one lumen is typically visible. When the scope reaches a branching point, 2 or more lumens are tracked. The observation likelihood $P(Z_t|k)$ of transitioning into a specific child node is adjusted according to whether each tracked lumen is active, and currently seen in the frame, or not. Once the number of active tracked lumens goes back to one, the likelihood strongly indicates that the scope has gone through the branching point and into the active tracked lumen. 

\noindent Similarly, for backward movement, if the detected lumens go from one in the original location to two or more active lumens, the observation likelihood strongly indicates that the scope has retracted into the parent node's branching point.

\subsection{Switch Verification.} \label{sec:svmodule}
Since lumen tracking can sporadically fail, the Switch Verification module verifies that a detected location change corresponds to a genuine transition rather than a sporadic detection or tracking error.
If $P_{loc,t}$ indicates a possible location change, meaning \[
\operatorname*{arg\,max}_{k \in S^{tree}} P_{loc,t}(k) \not = S_{t-1},
\] the Switch Verification module compares the current frame $I_{change}$ with the last frame $I_{last}$ with a different number of active tracked lumens:
\begin{itemize}
    \item If forward motion, the last frame with at least two active tracked lumens (two tracked lumens with $n_{lost}=0$).
    \item If backward motion, the last frame with one tracked lumen. 
\end{itemize}
The comparison is done using a lightweight siamese network consisting of three convolutional layers, and trained to compute similarity between frames with one and more than one lumen. If the difference between $I_{change}$ and $I_{last}$ is greater than a threshold (determined on a validation set of frames), the images are considered different enough and the location switch correct. If not, the system waits for $n_f$ frames to re-evaluate the scene, and the probability updates are put on hold. If after the $n_f$ frames the number of visible lumens is still the same than in $I_{change}$, the location change is considered valid. Otherwise, the location change is rejected and the previous location is maintained.
Therefore, the output of this module is the updated scope location probability if the location switch is valid, or the scope location probability from the previous frame, maintaining the same location, if the switch is not considered valid.
In our experiments, we empirically set $n_f = 10$ frames, approximately 0.3 seconds of video at 30fps. This value was chosen to be long enough to bridge temporary tracking losses caused by motion blur or specular reflections, but short enough to prevent noticeable navigational lag.

\section{BronchoTop dataset}
\label{sec:dataset}

Due to a lack of public annotated real data for this task, we collect and release a new benchmark for topological localization in bronchoscopy data, the BronchoTop dataset. It is a unique set that contains real and simulated bronchoscopy sequences with frame-level topological location labels provided by experts for each frame. It consists of three subsets: \textbf{\textit{BronchoTop-Det}}, \textbf{\textit{BronchoTop-Sim}} and \textbf{\textit{BronchoTop-Real}}, summarized in Table~\ref{tab:data_ours}. 
Frames from all our subsets have resolution 256x256. Figure~\ref{fig:dataset} shows sample frames from each released set. 
\begin{itemize}
    \item \textbf{BronchoTop-Sim} (Simulated) set: 9 simulated (\textit{Sim1}-\textit{Sim9}) sequences generated from CT-based 3D airway reconstructions of 2 patients, each containing 700 to 1000 frames. Sequences follow different paths down to branching level 4 (e.g., nodes TriLLB or LB4+5 in the airway tree model). Camera can only go forward, towards deeper levels.
    Sequences are generated using \texttt{Blender}, and the framework is made available to generate additional sequences.
    
    \item \textbf{BronchoTop-Real} set: 6 real (\textit{Real1}-\textit{Real6}) bronchoscopy sequences recorded during real procedures on 3 different patients, containing 505 to 1285 frames each. Sequences follow different paths down to branching level 3 (e.g., nodes BI or LLB6 in the airway tree model). Camera motion includes both forward-only and combined forward–backward trajectories. Sequences \textit{Real1} to \textit{Real4} have been labeled manually by doctors during the procedure, with one topological location label from the bronchial tree model assigned to each frame. 
    Sequences \textit{Real5} and \textit{Real6} do not have corresponding topological location labels. 

    \item \textbf{BronchoTop-Det} (Detection) set: a limited set of 90 bronchoscopy images with manually defined ground-truth lumen detection bounding-boxes. Images are selected from sequences \textit{Real1} and \textit{Real2} of set \textit{BronchoTop-Real}, with varying illumination and visibility conditions. 
\end{itemize}

\begin{table}[tb]
    \centering
    \setlength\tabcolsep{4pt}
    \caption{\textbf{Proposed BronchoTop dataset}, divided into three subsets of simulated (\textbf{Sim}) and real data, for Lumen Detection (\textbf{LD}) and Scope Topological Localization (\textbf{Loc.}) into anatomical regions. 
    }
    \label{tab:data_ours}
    \begin{tabular}{cccclc}
         \textbf{Dataset} & \textbf{Type} & \textbf{Task} & \textbf{Sequences} & \textbf{Labeled}  & \textbf{Split} \\
         \toprule
         BronchoTop-Det & Real & LD & * & \checkmark (manual) & Test \\
         \midrule
         \multirow{2}{*}{BronchoTop-Sim} & \multirow{2}{*}{Sim.} &  \multirow{2}{*}{Loc.} & \textit{Sim1} - \textit{Sim7} & \checkmark (auto) & Test \\
          & & & \textit{Sim8}, \textit{Sim9} & \checkmark (auto) & Train \\
         \midrule
         \multirow{2}{*}{BronchoTop-Real} & \multirow{2}{*}{Real} &  \multirow{2}{*}{Loc.} & \textit{Real1} - \textit{Real4} & \checkmark (manual) & Test \\
          & & & \textit{Real5}, \textit{Real6} & \ding{55} & Train \\
         \bottomrule
         \multicolumn{6}{p{8cm}}{*90 diverse frames, manually annotated, from sequences Real1 and Real2 of dataset \textit{BronchoTop-Real}.}\\
    \end{tabular}  
\end{table}

\begin{figure}[tb]
    \centering
    \setlength\tabcolsep{2pt}
    \begin{tabular}{ccc ccc}
       \includegraphics[width=0.15\linewidth]{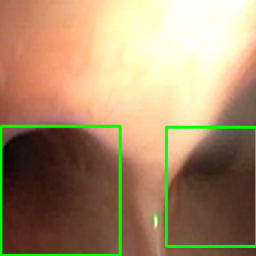} & \includegraphics[width=0.15\linewidth]{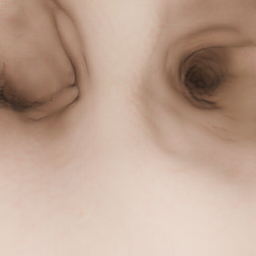} & \includegraphics[width=0.15\linewidth]{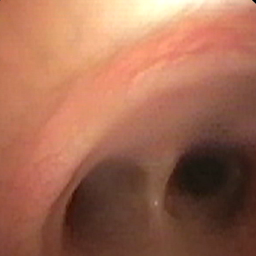} &
       \includegraphics[width=0.15\linewidth]{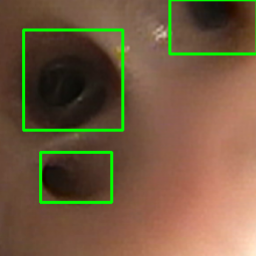} & \includegraphics[width=0.15\linewidth]{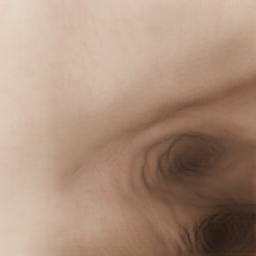}& \includegraphics[width=0.15\linewidth]{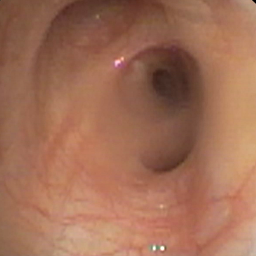}\\
        (a) & (b) & (c) & (a) & (b) & (c)\\
    \end{tabular}
    \caption{\textbf{Proposed BronchoTop dataset.} Examples of frames from sets (a) BronchoTop-Det with corresponding ground-truth lumen detection bounding boxes (in green) (b) BronchoTop-Sim, (c) BronchoTop-Real.}
    \label{fig:dataset}
\end{figure}

\section{Experiments and Results}
This section presents the evaluation of BronchoTop on phantom, simulated, and real bronchoscopy sequences. 
To incorporate reference results, we have implemented completely or partially (if the whole approach was not replicable) the most relevant related works. Since we could not find original implementations, we implemented the most relevant baselines found, with the required adaptations to run the online topological localization task on the public datasets considered.

Our approach includes both original ingredients and modules that are common in existing related work pipelines (\textit{Lumen detection and Tracking}, module A, and \textit{Scope Location Probability Estimation}, module C). We individually analyze the performance of our proposed implementation for modules A and C with respect to existing alternatives in the literature (experiments in Sections~\ref{sec:res_det} and Section~\ref{sec:moduleC}).
The validation of the steps that are specific to our pipeline (algorithm for \textit{Lumen-Branch Association}, module B, and \textit{Switch Verification}, module D) is run on the main experiment (Section~\ref{sec:loc}). It compares the performance of our complete BronchoTop approach with the closest related work on topological localization in bronchoscopy. 

\subsection{Data}\label{sec:data}
We evaluate our approach using a combination of public and newly collected datasets. In addition to two existing public benchmarks, we use our new BronchoTop dataset.  
Table~\ref{tab:data} summarizes the different datasets used for each task.

\paragraph*{Lumen detection task} the lumen detection module is evaluated on two datasets, our new \textbf{BronchoTop-Det} data (details in Sec.~\ref{sec:dataset}) and the publicly available \textbf{CVC-DBLumen}~\cite{sanchez2013line}  (125 real bronchoscopy frames along with ground-truth lumen segmentation masks. Ground-truth lumen detection bounding boxes are obtained from the binary segmentation masks by grouping pixels into connected components). 
In both sets, each image can contain multiple lumens. 

\paragraph*{Topological localization task}
the performance of BronchoTop is evaluated on simulated and real bronchoscopy videos from our \textbf{BronchoTop-Sim} and \textbf{BronchoTop-Real} data (details in Sec.~\ref{sec:dataset}) and a publicly available \textbf{phantom dataset}~\cite{visentini2017deep}. These phantom sequences were obtained by manually navigating a camera through a physical phantom model of the airway. The phantom follows the same airway structure as our bronchial tree model in Figure\ref{fig:intro}. The sequences used for evaluation in this work, \textit{Phantom0} and \textit{Phantom7}, as well as the ground-truth labels, are the same as used in~\cite{keuth2024airway}. Similarly, the other 14 sequences in the Phantom dataset are used for training.

\begin{table}[tb]
    \centering
    \setlength\tabcolsep{5pt}
    \caption{Summary of datasets used for Lumen Detection and Scope Topological Localization Tasks.}
    \label{tab:data}
    \begin{tabular}{lccc}
        \textbf{Dataset} & \textbf{Type} & \textbf{Task} & \textbf{Annotation*} \\ 
        \toprule
        CVC-DBLumen~\cite{sanchez2013line} & Real & Lumen Detection & SM/BB  \\
        BronchoTop-Det & Real & Lumen Detection &  BB  \\
        \midrule
        Phantom~\cite{visentini2017deep} & Phantom & Localization & TL \\
        BronchoTop-Sim & Simulated & Localization & TL \\
        BronchoTop-Real & Real & Localization & TL  \\
        \bottomrule
        \multicolumn{4}{p{8cm}}{* Type of annotation:  \textbf{SM}: segmentation mask; \textbf{BB}: bounding box; \textbf{TL}: topological location}\\
    \end{tabular}
\end{table}


\subsection{Lumen detection evaluation (module A).} \label{sec:res_det}

This experiment evaluates the performance of the proposed lumen detection method (module A in our approach) compared to existing strategies proposed in the literature. 

\begin{itemize}
    \item Method \textbf{M1}: As described in~\cite{sanchez2013line}, lumens are detected using image gradients and pixel intensity. The original method is evaluated on CVC-DBLumen dataset. 
    \item Method \textbf{M2}: Based on~\cite{wang2021depth}, a depth map is generated for each image using CycleGAN trained on simulated images. The depth map is then thresholded to keep only the darkest pixels, as lumens are the furthest points from the camera and correspond to local maxima in the depth map.
    \item Method \textbf{M3}: Following~\cite{keuth2023weakly}, we train Lite R-ASPP~\cite{howard2019searching} segmentation model on simulated data with pseudo-labels. To obtain pseudo-labels for training, depth maps are generated along with the simulated images, then segmented using K-means clustering with $K=2$.
\end{itemize}
We re-implement methods M2 and M3. For method M1, no code is available and the training set is not public, so re-implementation is not possible. Then, results for M1 are those reported by the authors~\cite{sanchez2013line} on the original evaluation dataset. 

Table~\ref{tab:detection} presents precision and recall for lumen detection using each method on datasets CVC-DBLumen and BronchoTop-Det. The results show that our detection method achieves the best balance between precision and recall results on both datasets, while maintaining low execution time, 17ms per image, enabling real-time inference. While method M1 performs better on CVC-DBLumen dataset, its inference time, at 57ms, does not allow for real-time running of the pipeline, around 30ms per image. Additionally, the lower precision of our method on individual frames is mitigated downstream. The sporadic false positives are filtered out by our temporal tracking and switch verification modules.


    

\begin{table}[tb]
\centering
\scriptsize 
\setlength\tabcolsep{3pt} 
\caption{\textbf{Lumen Detection Performance Comparison}.}
\label{tab:detection}

\begin{tabularx}{\columnwidth}{l @{\hspace{10pt}} ccc @{\hspace{10pt}} ccc}
\toprule
 & \multicolumn{3}{c}{\textbf{CVC-DBLumen \cite{sanchez2013line}}} & \multicolumn{3}{c}{\textbf{BronchoTop-Det}} \\
\cmidrule(lr){2-4} \cmidrule(lr){5-7}
\textbf{Method} & \textbf{Prec.(\%)} & \textbf{Rec.(\%)} & \textbf{Time (ms)} & \textbf{Prec.(\%)} & \textbf{Rec.(\%)} & \textbf{Time (ms)} \\
\midrule
M1 \cite{sanchez2013line}$^*$ & 98 & 97 & 57 & $-$ & $-$ & 57 \\
M2 \cite{wang2021depth}$^+$ & 85 & 84 & 5  & 76 & 81 & 5  \\
M3 \cite{keuth2023weakly}$^+$ & 66 & 62 & 14 & 82 & 65 & 14 \\
\textbf{Ours} & \textbf{80} & \textbf{91} & 17 & \textbf{90} & \textbf{88} & 17 \\
\bottomrule

\addlinespace
\multicolumn{7}{l}{\footnotesize $^*$: Results as reported by authors in \cite{sanchez2013line}.}\\
\multicolumn{7}{l}{\footnotesize $^+$: Results with our implementation of the approach.}\\
\multicolumn{7}{l}{\footnotesize $-$: No code available for evaluation on dataset BronchoTop-Det.}
\end{tabularx}
\end{table}

\subsection{Scope location probability estimation evaluation (module C).} \label{sec:moduleC}

As discussed, several existing approaches follow similar structure than ours (lumen detection, tracking and localization estimation)~\cite{sanchez2016navigation, wang2022bronchial, tian2024bronchotrack}. Since we could not replicate the initial lumen detection and tracking stages, due to a lack of algorithmic details or training data to replicate the proposed models, we use the same detection and tracking modules than our approach (modules A and B, detailed in Sec.~\ref{sec:method}), but then replace the final scope localization estimation step (module C) with their proposed strategy.

\begin{itemize}
    \item  Sanchez et al. ~\cite{sanchez2016navigation} is used to implement Baseline \textbf{B1}. The location of the bronchoscope is determined for each frame based on the number of tracked lumens in the previous and current frames: the scope moves to a lower level in the tree when the number of tracked lumens in the current frame is strictly lower than the number of tracked lumens in the previous frame. 

    \item Wang et al.~\cite{wang2022bronchial} is followed to implement Baseline \textbf{B2}. The location is estimated using the camera moving direction and the tracked lumens. If all tracks are matched or there are unmatched tracks in both previous and current frames, the location is the same. Otherwise, unmatched lumens in previous or current frame determine whether the scope has moved forward or backward along the tree.
    
    \item Tian et al.~\cite{tian2024bronchotrack} is followed to implement Baseline \textbf{B3}. The location depends on the tracked lumen labels and the known lumens theoretically visible at each node in the bronchial tree model. 

\end{itemize}

\begin{figure}[tb]
    \centering
    \includegraphics[width=0.9\linewidth]{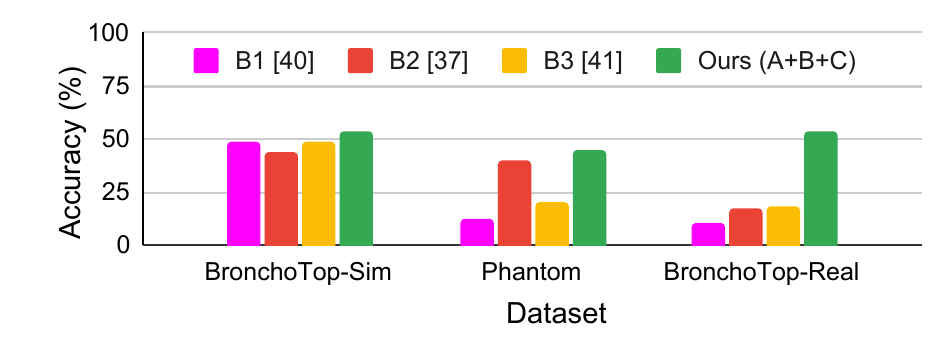}
\caption{\textbf{Scope Location Probability Estimation module}. Evaluation of different baselines and our approach. All approaches share the steps A and B (Lumen detection, tracking and labeling) and follow different baselines for the Scope location prediction (module C). Average accuracy reported on the three evaluation datasets (\textit{BronchoTop-Sim., BronchoTop-Real, Phantom}). 
}
\label{fig:chart_loc}
\end{figure}

This experiment, summarized in Figure~\ref{fig:chart_loc}, runs steps A, B and C of our pipeline varying the last step (C), that is performing the scope localization step. 
The average accuracy obtained by BronchoTop alternative,  our proposed module C, significantly outperforms those based on existing work. The benefits are particularly remarkable as the dataset used gets more challenging. 

\subsection{Topological Localization performance.} \label{sec:loc}

This experiment compares the topological localization results on all benchmarks for our full approach and the alternative strategy implemented as \textbf{B4}, following ~\cite{keuth2024airway}, the more complete alternative found that we could replicate based on existing related work. \cite{keuth2024airway} trains a ResNet9 model on phantom training data~\cite{visentini2017deep} and combines it with a Bayesian temporal filter to compute the probability of each frame belonging to a node in the bronchial tree model based on the previous frames' location. For fair comparison, we replace the original offline temporal filter (applied once the full sequence is available) with an online filter (ran as each new frame arrives).
The metric used in the following experiment is topological localization accuracy, defined as the percentage of total frames where the system's predicted anatomical node matches the expert-annotated ground-truth, meaning nodes must be exactly identified, such as RMB or LLB6.




Table~\ref{tab:sv_res} details the localization accuracy results of our method with and without Switch Verification module, against baseline B4. 
Both BronchoTop variants significantly outperform B4 on BronchoTop-Sim and BronchoTop-Real. Although B4 achieves higher accuracy on the Phantom~\cite{visentini2017deep} dataset, this highlights a critical limitation of this baseline: its CNN classification module needs to be trained supervised on each target dataset. However, labels are expensive to obtain, and currently available only for Phantom data. Trained on Phantom data, it drastically fails to generalize to real data. In contrast, BronchoTop only requires labels to train the similarity siamese network, which are much easier to obtain.
Also, note that, while the Phantom train and test sets consist of different sequences, these sequences were captured using the same phantom airway, with the same appearance characteristics, as if it would correspond to a single patient, and B4 might overfit to this specific physical airway.

These results also highlight the significant improvements brought to our approach by incorporating the proposed Switch Verification (SV) module. When comparing our full pipeline BronchoTop (A+B+C+D) to its ablated version without the SV module (A+B+C), we observe an increase of 21\% and comparable results across sequences. 
These results indicate that the SV module is crucial for verifying suspected location changes before updating the scope’s position, and achieving robust, real-time bronchoscopy navigation.

\begin{table}[tb]
\centering
\footnotesize 
\setlength\tabcolsep{2.5pt} 
\caption{\textbf{Topological localization results}. Accuracy reported for each sequence in datasets \textit{Phantom (Ph.)}, \textit{Sim} and \textit{Real}, for methods \textbf{B4} (adapted from~\cite{keuth2024airway}) and our approach \textbf{BronchoTop} with and without the SV Module (D). }
\label{tab:sv_res}
\begin{tabular}{lccc|lccc}
\toprule
\textbf{Sequence} & \textbf{BT} & \textbf{BT+D} & \textbf{B4} & \textbf{Sequence} & \textbf{BT} & \textbf{BT+D} & \textbf{B4} \\
\midrule
Sim1 & \textbf{93.24} & 90.31 & 28.19 & Ph.0 & 58.72 & 68.42 & \textbf{77.86} \\
Sim2 & 19.64 & \textbf{87.86} & 29.12 & Ph.7 & 30.63 & 35.34 & \textbf{61.88} \\
Sim3 & 82.92 & \textbf{92.88} & 29.42 & \cellcolor{lightgray} \textit{Ph. Avg} & \cellcolor{lightgray} 44.67 & \cellcolor{lightgray} 51.88 & \cellcolor{lightgray} \textbf{69.87} \\
Sim4 & 52.40 & \textbf{88.01} & 11.42 & 
Real1 & 73.27 & \textbf{82.38} & 13.47 \\
Sim5 & 28.44 & \textbf{81.44} & 15.80 & Real2 & 43.64 & \textbf{92.59} & 36.23 \\
Sim6 & 81.64 & \textbf{82.63} & 38.20 & Real3 & 25.37 & \textbf{59.17} & 24.59 \\
Sim7 & 18.56 & \textbf{96.41} & 50.80 & Real4 & 73.28 & \textbf{75.78} & 10.01 \\
\cellcolor{lightgray} \textit{Sim. Avg} & \cellcolor{lightgray}53.83 & \cellcolor{lightgray}\textbf{88.51} & \cellcolor{lightgray}28.99 & \cellcolor{lightgray}\textit{Real. Avg} & \cellcolor{lightgray}53.89 & \cellcolor{lightgray}\textbf{77.48} & \cellcolor{lightgray}21.07 \\
\bottomrule
\multicolumn{8}{l}{\footnotesize \textbf{BT}: BronchoTop (A+B+C), \textbf{BT+D}: BronchoTop (A+B+C+D)}\\
\end{tabular}
\end{table}

\begin{figure*}[tb]
    \centering
    \setlength\tabcolsep{1pt}
    \begin{tabular}{cc}
     \includegraphics[width=.49\linewidth]{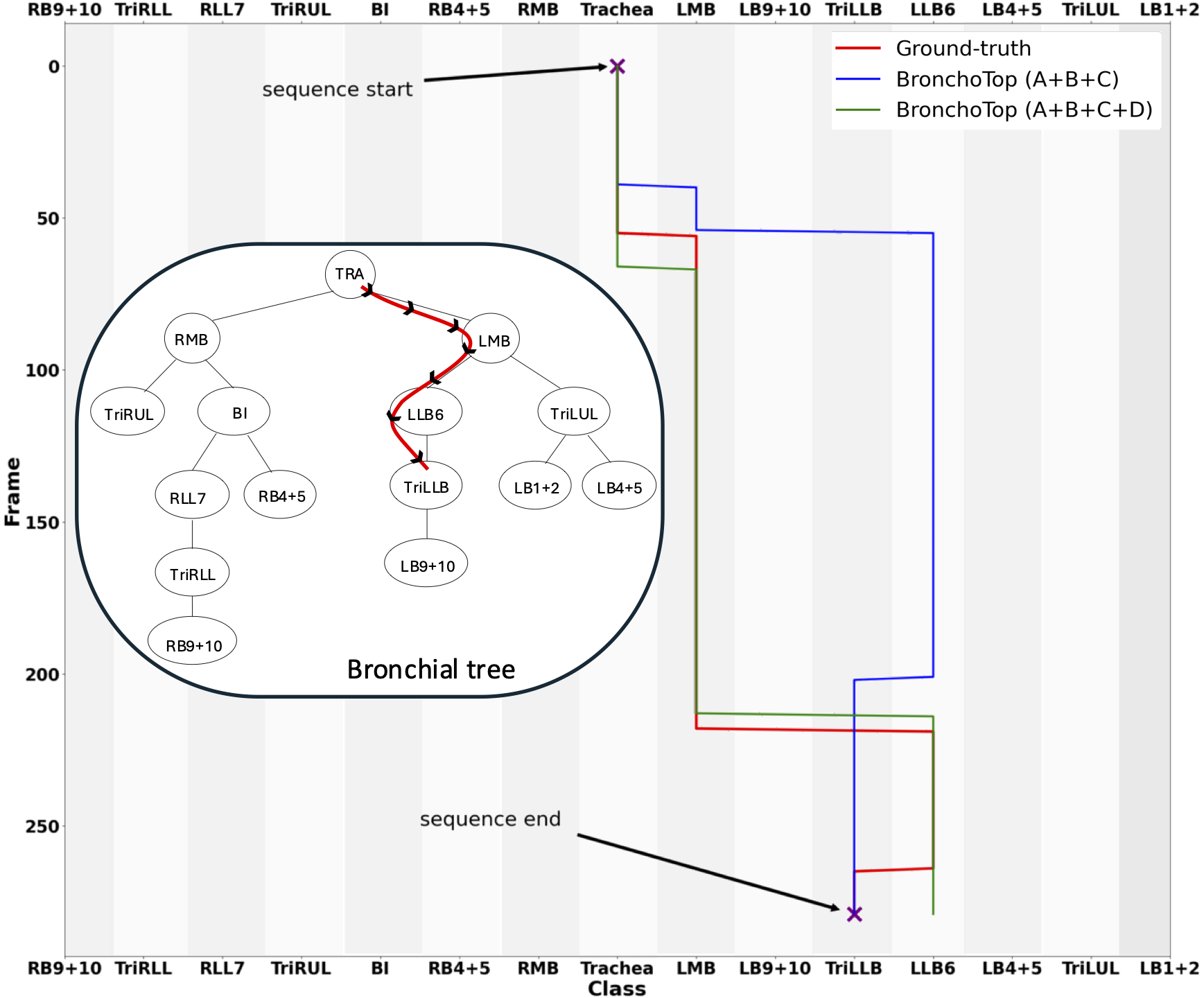} & \includegraphics[width=.49\linewidth]{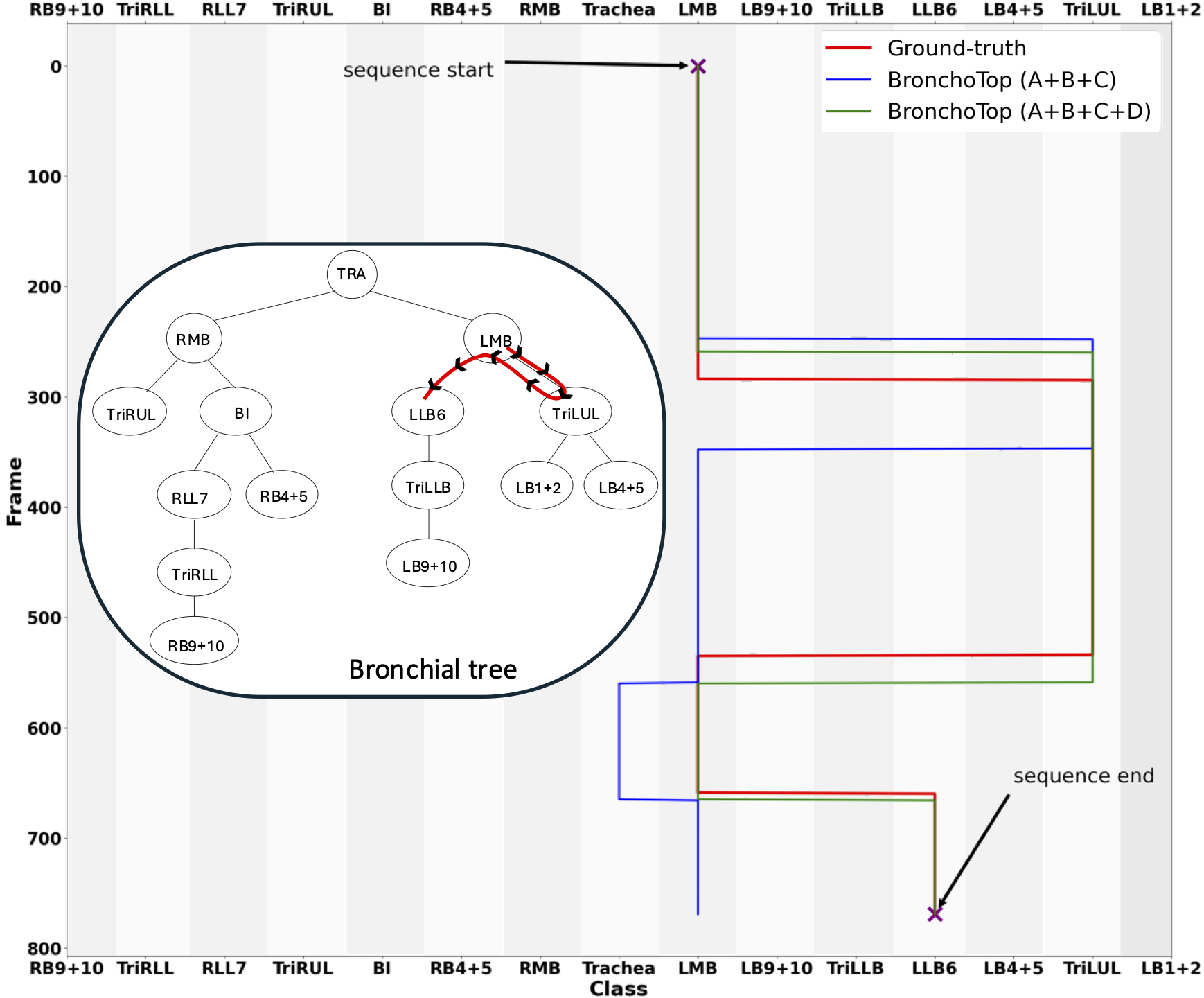}\\ 
     (a) Localization along sequence \textit{Sim2} &
     (b) Localization along sequence \textit{Real2}\\
    \end{tabular}
    
    \caption{\textbf{Scope location evolution} along sequences (a) Sim2 and (b) Real2. Location estimated using BronchoTop without module D (BronchoTop (A+B+C), in blue) and with module D (BronchoTop (A+B+C+D), in green), compared to ground-truth location (in red). Time progression shown from top to bottom.}
    \label{fig:qual_loc}
\end{figure*} 

Figure~\ref{fig:qual_loc} presents additional qualitative results, illustrating the progression of the estimated scope location across the bronchial tree along sequences \textit{Sim2} and \textit{Real2} using BronchoTop with or without the SV module. 
The diagrams show that the full BronchoTop approach is better adjusted to the ground-truth location than the simplified version of BronchoTop without module D. For example, in sequence Sim2, BronchoTop (A+B+C) switches from location LMB to location LLB6, which differs from the ground-truth, while the Switch Verification module is able to discard this potential switch and keep location LMB. Similarly, in sequence Real2, BronchoTop (A+B+C) switches from location LMB to location TRA while the Switch Verification module in BronchoTop discards this switch and maintains the correct location LMB.

These diagrams also highlight a limitation inherent to tracking-based methods. A single topological estimation error will propagate to subsequent predictions, even if future transitions are correctly detected. For example, in sequence Real2 (diagram (b)), BronchoTop (A+B+C) wrongly estimates location TRA instead of LMB. The next location switch detected is correct, but the location is estimated as LMB (one down from TRA) instead of LLB6. This would require manual correction by the clinician.

Finally, a supplementary video is provided showing the performance of BronchoTop on full sequences.

\section{Conclusions}
This work presented BronchoTop, the first real-time, RGB-only framework for topological localization in bronchoscopy. In contrast to existing navigation approaches that depend on patient-specific CT scans, additional sensors, or complex registration methods, BronchoTop leverages a generic bronchial tree model. Our four-stage pipeline, integrating lumen detection and tracking, lumen-branch label association, probabilistic scope location and a switch verification module, enables robust and efficient scope location estimation using clinical bronchoscopy video. Experimental results on phantom, simulated and real bronchoscopy sequences demonstrate that the proposed approach achieves state-of-the-art localization accuracy, with significant improvements over existing baselines, while maintaining real-time performance. Additionnally, we release the BronchoTop dataset, the first public benchmark for topological localization, enabling reproducible evaluation and further work on the topic.

Future work will focus on extending BronchoTop to deeper, more-complex levels of the airway, as well as improving robustness to incorrect location changes and their influence on subsequent estimated locations, and exploring integration of BronchoTop to clinical and robotic bronchoscopy systems. 


\bibliographystyle{IEEEtran}
\bibliography{bibliography}

\end{document}